\documentclass[sigconf]{acmart}

\usepackage{booktabs} 

\newfont{\mycrnotice}{ptmr8t at 7pt}
\newfont{\myconfname}{ptmri8t at 7pt}

\usepackage[ruled,vlined]{algorithm2e}
\usepackage{amsmath}
\usepackage{arydshln}

\setcopyright{rightsretained}

\begin{document}

\copyrightyear{2020}
\acmYear{2020}
\setcopyright{rightsretained}
\acmConference[GECCO '20 Companion]{Genetic and Evolutionary Computation Conference Companion}{July 8--12, 2020}{Cancún, Mexico}
\acmBooktitle{Genetic and Evolutionary Computation Conference Companion (GECCO '20 Companion), July 8--12, 2020, Cancún, Mexico}
\acmDOI{10.1145/3377929.3390038}
\acmISBN{978-1-4503-7127-8/20/07}

\title{Improving BPSO-based feature selection applied to offline WI handwritten signature verification through overfitting control}

\author{Victor L. F. Souza, Adriano L. I. Oliveira}
\affiliation{%
  \institution{Centro de Inform\'atica, UFPE}
  \city{Recife} 
  \state{Pernambuco} 
  \country{Brazil}
}
\email{vlfs@cin.ufpe.br, alio@cin.ufpe.br}


\author{Rafael M. O. Cruz, Robert Sabourin}
\affiliation{%
  \institution{\'Ecole de Technologie Sup\'erieure - Universit\'e du Qu\'ebec}
  \city{Montreal} 
  \state{Qu\'ebec} 
  \country{Canada}
}
\email{rafaelmenelau@gmail.com, robert.sabourin@etsmtl.ca}

\begin{abstract}
This paper investigates the presence of overfitting when using Binary Particle Swarm Optimization (BPSO) to perform the feature selection in a context of Handwritten Signature Verification (HSV). \textit{SigNet} is a state of the art Deep CNN model for feature representation in the HSV context and contains 2048 dimensions. Some of these dimensions may include redundant information in the dissimilarity representation space generated by the dichotomy transformation (DT) used by the writer-independent (WI) approach. The analysis is carried out on the GPDS-960 dataset. Experiments demonstrate that the proposed method is able to control overfitting during the search for the most discriminant representation.
\end{abstract}


%
%



\maketitle

\section{Introduction}


In the Writer-Independent Handwritten Signature Verification (WI-HSV) approach, a single model is trained from a dissimilarity space and is responsible for verifying the signatures of any available writer in the dataset. Thus, the classification inputs are dissimilarity vectors, which represent the difference between the features of a queried signature and a reference signature of the writer.

The \textit{SigNet}, proposed by \citet{hafemann:17}, is a state of the art Deep Convolutional Neural Network (DCNN) model for feature representation in the HSV context and its feature vectors are composed of 2048 dimensions. 
Some of the features may be redundant when transposed to a WI dissimilarity space.

Thus, we propose to use a feature selection technique based on binary particle swarm optimization (BPSO) to obtain only the relevant dimensions on this transposed space \cite{chuang:11}.
The optimization is conducted based on the minimization of the Equal Error Rate ($EER$) of the SVM in a wrapper mode \cite{radtke:06}. 
In this scenario, one possible problem that can be faced is overfitting. 
Thus, the objectives of this study are: (i) to analyze the redundancy in the features obtained in the dissimilarity space generated by DT. (ii) The presence of overfitting when using Binary Particle Swarm Optimization (BPSO) to perform the feature selection in a wrapper mode. (iii) Whether overfitting control can improve optimization performance.



\section{Basic concepts}
\label{sec:basic_concepts}

\paragraph{\textbf{Writer-Independent Handwritten Signature Verification \\(WI-HSV)}}


The Dichotomy Transformation (DT) allows to transform a multi-class pattern recognition problem into a $2$-class problem. 
In this approach, a dissimilarity (distance) measure is used to distinguish whether a given reference and a questioned sample belong to the same class or not \citep{souza:20}. When applied to the HSV context it characterizes the writer-independent (WI) approach, the samples are signatures and to perform the verification means belonging to the same writer (\textit{positive class}) or not (\textit{negative class}) \citep{souza:20}.  

The dissimilarity vector resulting from DT, $\textbf{u}$, is obtained by applying the absolute value of the difference from $x_{qi}$ and $x_{ri}$, where $i$ is the respective feature of the questioned signature signature ($\textbf{x}_q$) and reference signature ($\textbf{x}_r$) \citep{souza:20}.

\paragraph{\textbf{Feature selection using BPSO}}



In a context of feature selection, particle swarm optimization algorithms are used in their binary version (BPSO) and have been obtaining good results \cite{chuang:11}. 
We use a variation of PSO, in which the algorithm itself adjusts {\it w}, {\it $c_1$} and {\it $c_2$} dynamically over iterations, promoting global search in the beginning and local search in the final iterations \cite{zhang:13}.


The transformation of the continuous search space into a binary space is conducted by using a V-shaped transfer function \cite{mirjalili:13}.




We propose to use a BPSO-based feature selection for WI-HSV in a wrapper mode. 
The optimization is conducted based on the minimization of the Equal Error Rate ($EER$) of the SVM in a wrapper mode. The user threshold (considering just the genuine signatures and the skilled forgeries) was employed \cite{souza:20}.


In the feature selection scenario, overfitting occurs when the optimized feature set memorizes the training set instead of producing a general model. 
To decrease the chance of overfitting, a validation procedure can be used during the optimization process in order to select solutions with good generalization power.

According to \citet{radtke:06}, one possible validation strategy is the \textit{last iteration strategy}, this approach validate final candidate solutions on another set of unknown observations – the selection set. By using this approach, the optimization routine produces better results than selecting solutions based solely on the accuracy of the optimization set alone. However, this strategy has the disadvantage that the solution is validated only once, after the optimization process is completed.

Another approach is the \textit{global validation strategy} \cite{radtke:06}, where the validation of the candidate solutions are executed in all iterations of the optimization process. This can be accomplished by storing the best validated solutions in an external (auxiliary) archive.


During both validation routines, the optimization set ($Opt$) is temporarily replaced by the selection set ($Sel$) to evaluate the fitness function. 
In the \textit{global validation strategy}, at each iteration, all the best solutions previously found are grouped with the population of the new swarm and then ranked. Finally, the external archive maintains the $N$ best candidate solutions.




\section{Experiments}
\label{sec:experiments}


The experiments are carried out using GPDS-960 dataset, specifically in the GPDS-300 stratification. The Exploitation set, where the tested set is acquired, is composed of writers 1 to 300. The Development set is formed by the other 581 writers, from these: 146 writers are randomly selected to compose the train set, another 145 for the validation set, another 145 for the optimization set ($Opt$) and the remaining 145 for the selection set ($Sel$).


As in the work by \citet{souza:20}, we use the highest value for the number of references, i.e., 12 references per writer, and the Max function as the partial decisions. 
In the training step, the model uses 10 genuine signatures and 10 random forgeries. 
During optimization (optimization and selection sets), the fitness function minimizes the $EER$ with user threshold considering only genuine signatures and skilled forgeries. In this case, for each writer, 10 genuine signatures and 10 skilled forgeries are used.
These operations are performed in the space with reduced samples, i.e., after prototype selection through Condensed Nearest Neighbors (CNN) \citep{souza:20}.
The test set is acquired as in \cite{hafemann:17}. 
The SVM and IDPSO settings are the same as in \cite{souza:20} and \cite{zhang:13}, respectively. The maximum number of iterations was set to 40. Five replications were carried out.



\subsection{Results and discussions}


Table \ref{tab:models_comparison} presents the results obtained by the the models with and without feature selection. Table \ref{tab:state_gpds} contains the comparison of the presented models with the state of the art methods for the GPDS-300 dataset (references can be found in \cite{souza:20}).

\begin{table}[!htb]
\caption{Comparison of $EER$ considering the presented models, in the GPDS-300 dataset (errors in \%)}
\label{tab:models_comparison}
\footnotesize

\centering

\begin{tabular}{ccc}
\hline
Approach & \#features & $EER$ \\ 
\hline
No feature selection & 2048 & 3.57 (0.10) \\ 
Feature selection and no validation & 1124 & 3.76 (0.07) \\ 
Feature selection and last iteration validation & 1120 & 3.64 (0.08) \\ 
Feature selection and global validation & 1140 & \textbf{3.46 (0.08)} \\ 
 \hline
\end{tabular}
\end{table}


In terms of validation strategy, the improvement when using any of the validation stages was enough to obtain better results when compared to the model without feature selection.
Results indicate that not using a validation stage is worse than using validation at the last iteration, which in turn is worse than using the global validation strategy. 
Thus, by using the global validation strategy it is possible to control the overfitting of the model and, thereby, improve the performance of the BPSO-based feature selection approach.

Another aspect that can be observed is the presence of redundant features in the dissimilarity space generated by DT. Since, the model with feature selection and global validation uses only almost 55\% of the total number of features and still manages to obtain similar $EER$ when compared to the model trained with all the 2048 features.

\begin{table}[!htb]
\caption{Comparison of $EER$ with the state of the art, in the GPDS-300 dataset (errors in \%)}
\label{tab:state_gpds}
\scriptsize
\centering

\begin{tabular}{ccccc}
\hline
Type & HSV Approach & \#Ref & \#Models & $EER$ \\ 
\hline
WD & Hafemann et al. (2016)  & 12  &  300 & 12.83 \\ 
WD & Zois et al. (2016) & 5 & 300 &  5.48 \\ 
WD & Hafemann et al. (2017) & 5 & 300 & 3.92 (0.18) \\ 
WD & Hafemann et al. (2017) & 12 & 300 & 3.15 (0.18) \\ 
WD & Serdouk et al. (2017) & 10 & 300 &  9.30 \\ 
WD & Hafemann et al. (2018) & 12  & 300 & 3.15 (0.14) \\ 
WD & Hafemann et al. (2018) (fine-tuned) & 12  & 300 & 0.41 (0.05) \\ 
WD & Yilmaz and Ozturk (2018) & 12  & 300 & 0.88 (0.36) \\ 
WD & Zois et al. (2019) & 12 & 300 & 0.70 \\ 
WI & Dutta et al. (2016) & N/A & 1 &  11.21 \\ 
WI & Hamadene and Chibani (2016) & 5 & 1 &  18.42 \\ 
WI & Souza et al. (2019) & 12 & 1 & 3.47 (0.15) \\ 
WI & Zois et al. (2019) & 5 & 1 & 3.06 \\ 
\hdashline
WI & $SVM_{no-selection}$ & 12 & 1 & 3.57 (0.10) \\
WI & $SVM_{global-validation}$ & 12 & 1 & 3.46 (0.08) \\ 

\hline
\end{tabular}
\end{table}

In general, our $SVM_{global-validation}$ approach obtains low $EER$. In the WI scenario, it is only worse than the model proposed by Zois et al. (2019).
In the comparison with WD models, our approach only got worse results than Hafemann et al. (2018) (fine-tuned), Yilmaz and Ozturk (2018) and Zois et al. (2019), being better or comparable than the other cases.
It is important to point out that, as a WI model, our approach has greater scalability than these other models, since only one classifier is needed to perform signature verification.

\section{Conclusions}
\label{sec:conclusions}


In this work, we evaluated the use of BPSO-based feature selection for offline writer-independent handwritten signature verification. The optimization was conducted based on the minimization of the Equal Error Rate ($EER$) of the SVM in a wrapper mode. 
Experimental results showed that not using a validation stage is worse than using validation at the last iteration, which in turn is worse than using the global validation strategy. Thus, by using the global validation strategy it is possible to control the overfitting of the model and, thereby, improve the performance of the BPSO-based feature selection approach. 

\begin{acks}
  This work was supported by CNPq, FACEPE and \'ETS Montr\'eal. 
\end{acks}


\bibliographystyle{ACM-Reference-Format}
\bibliography{sample-bibliography} 


\begin{thebibliography}{6}


\ifx \showCODEN    \undefined \def \showCODEN     #1{\unskip}     \fi
\ifx \showDOI      \undefined \def \showDOI       #1{#1}\fi
\ifx \showISBNx    \undefined \def \showISBNx     #1{\unskip}     \fi
\ifx \showISBNxiii \undefined \def \showISBNxiii  #1{\unskip}     \fi
\ifx \showISSN     \undefined \def \showISSN      #1{\unskip}     \fi
\ifx \showLCCN     \undefined \def \showLCCN      #1{\unskip}     \fi
\ifx \shownote     \undefined \def \shownote      #1{#1}          \fi
\ifx \showarticletitle \undefined \def \showarticletitle #1{#1}   \fi
\ifx \showURL      \undefined \def \showURL       {\relax}        \fi
\providecommand\bibfield[2]{#2}
\providecommand\bibinfo[2]{#2}
\providecommand\natexlab[1]{#1}
\providecommand\showeprint[2][]{arXiv:#2}

\bibitem[\protect\citeauthoryear{Chuang, Tsai, and Yang}{Chuang
  et~al\mbox{.}}{2011}]%
        {chuang:11}
\bibfield{author}{\bibinfo{person}{Li-Yeh Chuang}, \bibinfo{person}{Sheng-Wei
  Tsai}, {and} \bibinfo{person}{Cheng-Hong Yang}.}
  \bibinfo{year}{2011}\natexlab{}.
\newblock \showarticletitle{Improved binary particle swarm optimization using
  catfish effect for feature selection}.
\newblock \bibinfo{journal}{{\em Expert Systems with Applications\/}}
  \bibinfo{volume}{38}, \bibinfo{number}{10} (\bibinfo{year}{2011}),
  \bibinfo{pages}{12699 -- 12707}.
\newblock
\showISSN{0957-4174}


\bibitem[\protect\citeauthoryear{Hafemann, Sabourin, and Oliveira}{Hafemann
  et~al\mbox{.}}{2017}]%
        {hafemann:17}
\bibfield{author}{\bibinfo{person}{Luiz~G. Hafemann}, \bibinfo{person}{Robert
  Sabourin}, {and} \bibinfo{person}{Luiz~S. Oliveira}.}
  \bibinfo{year}{2017}\natexlab{}.
\newblock \showarticletitle{Learning features for offline handwritten signature
  verification using deep convolutional neural networks}.
\newblock \bibinfo{journal}{{\em Pattern Recognition\/}}  \bibinfo{volume}{70}
  (\bibinfo{year}{2017}), \bibinfo{pages}{163 -- 176}.
\newblock
\showISSN{0031-3203}


\bibitem[\protect\citeauthoryear{Mirjalili and Lewis}{Mirjalili and
  Lewis}{2013}]%
        {mirjalili:13}
\bibfield{author}{\bibinfo{person}{Seyedali Mirjalili} {and}
  \bibinfo{person}{Andrew Lewis}.} \bibinfo{year}{2013}\natexlab{}.
\newblock \showarticletitle{S-shaped versus V-shaped transfer functions for
  binary Particle Swarm Optimization}.
\newblock \bibinfo{journal}{{\em Swarm and Evolutionary Computation\/}}
  \bibinfo{volume}{9} (\bibinfo{year}{2013}), \bibinfo{pages}{1 -- 14}.
\newblock
\showISSN{2210-6502}


\bibitem[\protect\citeauthoryear{{Radtke}, {Wong}, and {Sabourin}}{{Radtke}
  et~al\mbox{.}}{2006}]%
        {radtke:06}
\bibfield{author}{\bibinfo{person}{P.~V.~W. {Radtke}}, \bibinfo{person}{T.
  {Wong}}, {and} \bibinfo{person}{R. {Sabourin}}.}
  \bibinfo{year}{2006}\natexlab{}.
\newblock \showarticletitle{An Evaluation of Over-Fit Control Strategies for
  Multi-Objective Evolutionary Optimization}. In \bibinfo{booktitle}{{\em The
  2006 IEEE International Joint Conference on Neural Network Proceedings}}.
  \bibinfo{pages}{3327--3334}.
\newblock
\showISSN{2161-4407}


\bibitem[\protect\citeauthoryear{Souza, Oliveira, Cruz, and Sabourin}{Souza
  et~al\mbox{.}}{2020}]%
        {souza:20}
\bibfield{author}{\bibinfo{person}{Victor~L.F. Souza},
  \bibinfo{person}{Adriano~L.I. Oliveira}, \bibinfo{person}{Rafael~M.O. Cruz},
  {and} \bibinfo{person}{Robert Sabourin}.} \bibinfo{year}{2020}\natexlab{}.
\newblock \showarticletitle{A white-box analysis on the writer-independent
  dichotomy transformation applied to offline handwritten signature
  verification}.
\newblock \bibinfo{journal}{{\em Expert Syst Appl\/}} (\bibinfo{year}{2020}).
\newblock
\showISSN{0957-4174}


\bibitem[\protect\citeauthoryear{Zhang, Xiong, and Zhang}{Zhang
  et~al\mbox{.}}{2013}]%
        {zhang:13}
\bibfield{author}{\bibinfo{person}{YingChao Zhang}, \bibinfo{person}{Xiong
  Xiong}, {and} \bibinfo{person}{QiDong Zhang}.}
  \bibinfo{year}{2013}\natexlab{}.
\newblock \showarticletitle{An Improved Self-Adaptive PSO Algorithm with
  Detection Function for Multimodal Function Optimization Problems}.
\newblock \bibinfo{journal}{{\em Mathematical Problems in Engineering\/}}
  \bibinfo{volume}{vol. 2013} (\bibinfo{year}{2013}).
\newblock


\end{thebibliography}

\end{document}